\documentclass[journal,twoside]{IEEEtran}
\usepackage[ruled,linesnumbered]{algorithm2e}

\SetAlFnt{\small}
\SetAlCapFnt{\small}
\SetAlCapNameFnt{\small}
\SetAlCapHSkip{0pt}
\IncMargin{-\parindent}

\usepackage{url,subfigure,amsmath,amssymb,epsfig,verbatim,booktabs,graphicx,epstopdf}
\usepackage[colorlinks=false,linkcolor=black, citecolor=blue, urlcolor=black, pdfborder={0 1 0}]{hyperref}
\usepackage{cite}
\usepackage{bbm}
\usepackage{makecell}
\usepackage{mathrsfs}
\usepackage{mathtools}
\usepackage{etoolbox}
\usepackage{array}
\usepackage[justification=centering]{caption}
\usepackage{xcolor}

\begin{document}
\title{Heterogeneous Attentions for Solving Pickup and Delivery Problem via Deep Reinforcement Learning}
\author{
Jingwen Li$^{1}$, Liang Xin$^{2}$, Zhiguang Cao$^{1*}$, Andrew Lim$^{1}$, Wen Song$^{3}$, Jie Zhang$^{2}$

\thanks{$^{*}$ Corresponding Author.}
\thanks{$^{1}$Jingwen Li, Zhiguang Cao and Andrew Lim are with the Department of Industrial Systems Engineering and Management, National University of Singapore, Singapore (emails: lijingwen@u.nus.edu, zhiguangcao@outlook.com, isealim@nus.edu.sg).}

\thanks{$^{2}$Liang Xin and Jie Zhang are with the School of Computer Science and Engineering, Nanyang Technological University, Singapore (email: xinl0003@e.ntu.edu.sg, zhangj@ntu.edu.sg).}
\thanks{$^{3}$Wen Song is with the Institute of Marine Science and Technology, Shandong University, China (email: wensong@email.sdu.edu.cn).}
\thanks{This work is supported by the National Research Foundation of Singapore under grant NRF-RSS2016-004, the National Natural Science Foundation of China (Grant No. 61803104),  the Young Scholar Future Plan of Shandong University (Grant No. 62420089964188), and the Singtel-NTU Cognitive \& Artificial Intelligence Joint Lab through the NRF corporate lab@university scheme.} 
}

\markboth{
IEEE TRANSACTIONS ON INTELLIGENT TRANSPORTATION SYSTEMS, 2021}%
{~}

\maketitle
\begin{abstract}
Recently, there is an emerging trend to apply deep reinforcement learning to solve the vehicle routing problem (VRP), where a learnt policy governs the selection of next node for visiting. However, existing methods could not handle well the pairing and precedence relationships in the pickup and delivery problem (PDP), which is a representative variant of VRP. To address this challenging issue, we leverage a novel neural network integrated with a heterogeneous attention mechanism to empower the policy in deep reinforcement learning to automatically select the nodes. In particular, the heterogeneous attention mechanism specifically prescribes attentions for each role of the nodes while taking into account the precedence constraint, i.e., the pickup node must precede the pairing delivery node. Further integrated with a masking scheme, the learnt policy is expected to find higher-quality solutions for solving PDP. Extensive experimental results show that our method outperforms the state-of-the-art heuristic and deep learning model, respectively, and generalizes well to different distributions and problem sizes.

\end{abstract}

\begin{IEEEkeywords}
Heterogeneous attention, deep reinforcement learning, pickup and delivery problem
\end{IEEEkeywords}

\section{Introduction}
\label{sec:intro}

\IEEEPARstart{V}{ehicle} routing problem (VRP) is a fundamental topic in both communities of ITS (Intelligent Transportation Systems) and Operations Research, which has ubiquitous applications in industry, such as the logistics for harbor port~\cite{agra2015maritime}, airport~\cite{tang2015exact} and the general warehouse~\cite{anily1990one}. 
Instead of sharing a depot for all customers as VRP, in reality, a customer may always have his own delivery point, such as in intra-city express service~\cite{renaud2000heuristic} and  ride-sharing~\cite{agatz2012optimization}. The route planning for all applications of this type could be naturally described as a~\emph{pickup and delivery problem} (PDP), which is a representative variant of VRP. Generally, the PDP is characterized by pairing and precedence relationships, in which a pickup point must precede the pairing delivery point. 
Although widely studied, it is still challenging for conventional methods including exact and heuristic algorithms to optimally solve PDP with short computation time due to its NP-hard nature. Recently, there is an increasing attention on applying deep reinforcement learning (DRL) to automatically learn the rules in conventional heuristic methods for solving routing problems including Travelling Salesman Problem (TSP) and Capacitated VRP (CVRP), which delivers appealing results with much faster computation~\cite{bello2017neural,nazari2018reinforcement,kool2018attention,xin2021multi}. Inspired by them, in this paper, we aim to solve PDP by exploring deep reinforcement learning.

Although some success has been achieved, most of the DRL based solutions are only able to handle the typical VRP with a shared delivery point, which is less effective for coping with the pairing and precedence relationships in PDP. Intuitively, the masking scheme in existing DRL models could be adapted to represent those relationships in PDP, where the delivery points should be always masked until the pairing pickup points are visited. However, this adaption might be far from enough given the following two issues: 1) The masking scheme only takes effects in the output layer of the policy network, and a more desirable solution should enable the main body of the deep architecture intrinsically aware of the pairing and precedence relationships; 2) Different from the typical VRP, the nodes in PDP have more heterogeneous roles, such as ego pickup point, ego delivery point, peer pickup point, peer delivery point and depot. The complex relationship among them may render the decision making for selecting the next node more arduous.   

To cope with those challenging issues, in this paper, we propose a deep reinforcement learning based method that integrates with a heterogeneous attention mechanism for solving PDP. In particular, the policy network in the DRL is characterized by an encoder-decoder structure, and it learns constructing a solution by iteratively selecting a pickup or delivery point at each step. Regarding the heterogeneous attention mechanism, six types of attentions are specifically designed in addition to the original one. Among them, three types of attentions are used to learn the relationship between each ego pickup point to the points of other roles, and the remaining three types are used for each ego delivery point. Together with a masking scheme that adaptively masks out the invalid points to guarantee feasibility, the heterogeneous attention mechanism is supposed to empower the policy network to more intrinsically perceive and learn the pairing and precedence relationships. Extensive experimental results show that the proposed method outperforms both the state-of-the-art deep reinforcement learning method and conventional heuristic methods with much higher solution quality and shorter computation time. More importantly, the learned policy generalizes well to problems with different distributions and sizes. 

The remainder of the paper is organized as follows. Section \ref{sec:related work} briefly reviews conventional methods and deep models for routing problems. Section \ref{sec:problem descriptiton} introduces the PDP with mathematical formulation. Section \ref{sec:method} reformulates the problem in RL manner and elaborates our DRL solution. Section \ref{sec:experiments} provides the computational experiments and analysis. Finally, Section \ref{sec:conclusion} concludes the paper and presents future works.

\section{Related Work}
\label{sec:related work}
In this section, we review the related works on conventional methods for PDP, and (deep) learning models for routing problems. 

\subsection{Exact and Heuristic Methods}

Most of exact methods for solving PDP adopt branch-and-bound or its variants as framework. The precedence relations was first considered and added into classical TSP in~\cite{lokin1979procedures}, where an exact brand-and-bound method was adopted to tackle the problem. 
The general PDP was first surveyed in~\cite{savelsbergh1995general}, summarizing different variants of PDP with formulations and several exact methods such as brand-and-bound with additive bounding procedure and column generation scheme to handle these variants. One of the conclusions is that dynamic programming performed well on Single-vehicle Pickup and Delivery Problem (PDP) in small scale instances. A branch-and-cut algorithm was presented in~\cite{ruland1997pickup} to solve PDP based on three constraints: subtour elimination and precedence constraints, generalized order constraints and order matching constraints. 
A new branch-and-cut-and-price algorithm was proposed to solve PDP in~\cite{ropke2009branch} by considering two subproblems of pricing in column generation. However, despite delivering optimal solutions, the exact methods suffer from prohibitively heavy computation for large-scale ones given the exponential complexity.

In contrast, heuristic methods for PDP and VRP are able to tackle large-scale problems with relatively high computation efficiency~\cite{kim2016solving,cao2016improving,ghorai2020spea,cao2020using}. Based on the shift, exchange and rearrange operators, a tabu-embedded simulated annealing algorithm was proposed to cope with large-scale PDP in~\cite{li2003metaheuristic}.
An adaptive large neighborhood search heuristic method was presented in~\cite{ropke2006unified} to handle PDP by incorporating regret insertion method and six removal strategies such as shaw removal, cluster removal, etc.
Another adaptive large neighborhood search method was proposed in~\cite{ghilas2016adaptive} based on destroy and re-creat strategies. Starting from an initial solution generated by a greedy insertion heuristic, various efficient removal operators and insert principles were applied to improve the solution iteratively. A hybrid three-stage heuristic approach was presented to solve PDTSP in~\cite{hernandez2016hybrid}. With initial solutions generated using several local search operators, multi-start and variable neighborhood decent heuristic methods were integrated to improve the solution, where three shaking procedures were introduced to perturb solutions from local minimum. However, despite the desirable performance, the hand-engineered rules in the conventional heuristic methods heavily depend on human expertise and experience, which leave much room to improve the solution quality further.

\subsection{Learning Based Methods}
Recently, there is a growing trend to apply deep reinforcement learning to solve VRPs, and most of them are characterized by a policy network with an encoder-decoder structure. Generally, the encoder maps the 2-dimensional locations of customers (or nodes) into feature embedding to extract useful information from data, and the decoder addresses the problem in two different fashions, i.e., \emph{construction} or \emph{improvement}, respectively. Regarding the former, the decoder starts with an empty sequence, and iteratively selects a node at each step to construct a complete solution. Regarding the latter, the decoder starts with a complete initial solution, and constantly selects either candidate nodes or heuristic operators for certain operation at each step, so that the new solution could be improved over previous ones until the termination criterion is met. To force that each customer is only visited once, a masking scheme is always applied to mask the visited or invalid nodes. Further integrated with attention mechanisms or graph neural networks, those DRL models are potentially empowered to produce higher-quality solutions.

In specific, the first seminal deep architecture proposed to cope with routing problems was Pointer Network (PtrNet), which solved TSP in a supervised manner~\cite{vinyals2015pointer} and was later extended to the framework of reinforcement learning~\cite{bello2017neural}. Furthermore, PtrNet was adopted to solve CVRP based on the reinforcement learning, where the customer information (i.e., location and demands) was dynamic and the route length was uncertain~\cite{nazari2018reinforcement}. To speed up training process, the Recurrence Neural Network (RNN) structure in the encoder was removed since sequential and positional information was not meaningful for CVRP. A Transformer-based architecture was proposed in~\cite{kool2018attention,xin2020step} by adopting self-attention instead of Seq2Seq structure in both encoder and decoder to engender higher-quality solutions. A hybrid of local search and deep reinforcement learning was introduced in~\cite{zhao2020hybrid} to solve both VRP and VRPTW (time window). The reinforcement learning was adopted to solve the electric vehicle fleet dispatch problem in~\cite{shi2019operating}, in which a novel framework was proposed using the decentralized learning and centralized decision making. With the online routing problem transformed into a vehicle tour generation problem, a deep reinforcement learning based method was proposed using a structural graph embedded pointer network to construct those tours~\cite{james2019online}.
Rather than learning construction heuristics as the methods mentioned above, the NeuRewriter in ~\cite{chen2019learning} and the improvement heuristic method in~\cite{wu2020learning} were proposed to learn refining an initial yet complete solution iteratively with local search operators.

\begin{figure*}
\centering 
     \setlength{\abovecaptionskip}{0.1cm}
 	\setlength{\belowcaptionskip}{-0.5cm}
	\includegraphics[width=0.96\textwidth, height=58mm, trim={1mm 0mm 1mm 0mm},clip]{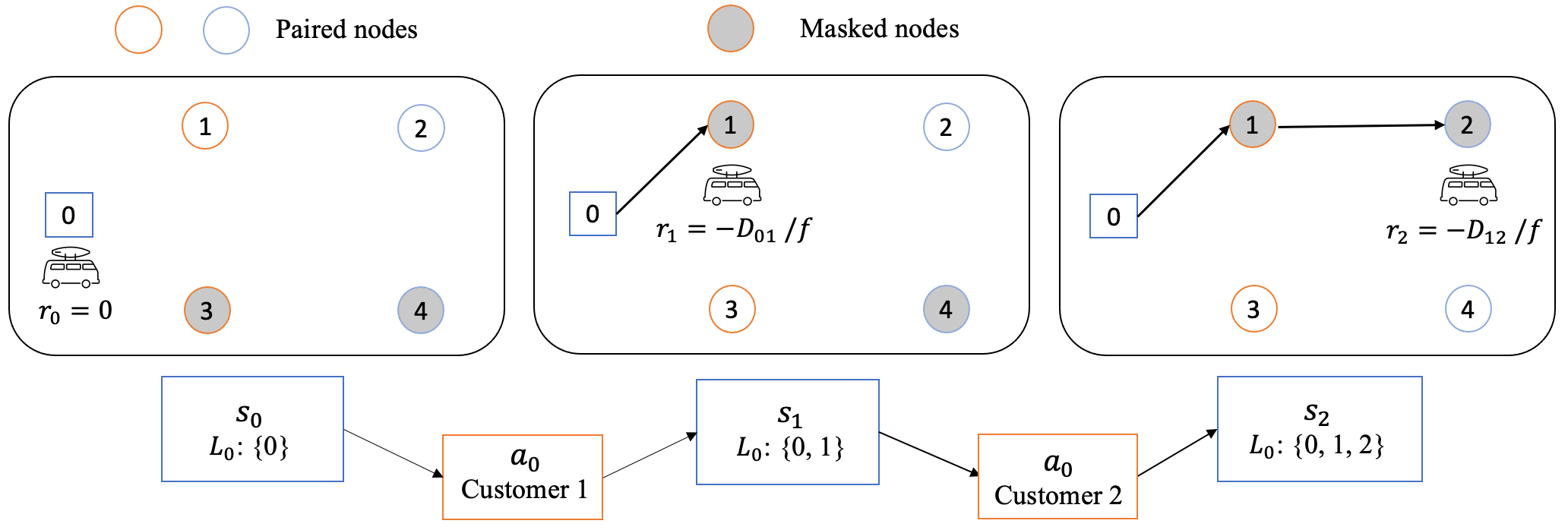} 
	\captionsetup{justification=justified}
	\caption{
	An illustrative example of MDP with 1 vehicle and 4 nodes. Node $x_1$ and $x_2$ are pickup nodes, and node $x_3$ and $x_4$ are their delivery nodes, where node $x_1$ and $x_3$, node $x_2$ and $x_4$ are paired. At the beginning, node $x_3$ and node $x_4$ are masked due to the precedence constraint. At the first step, node $x_1$ is selected, i.e., $a_1 = x_1$. In response to this action, a) node $x_1$ is masked and node $x_3$ becomes unmasked; b) the partial route is updated, i.e., $L_1=\{0, 1\}$; c) the accumulative reward for this step is the negative value of the travel time of the vehicle travelling from the depot to node $x_1$, i.e., $r_1 = - D_{01} / f$. At the second step, node $x_2$ is selected, and the state, action, reward are updated similarly.
	}
	\label{fig:MDP} 
\end{figure*}

\section{Preliminary}
\label{sec:problem descriptiton} 

In this section, we introduce the preliminary for PDP and describe its mathematical formulation. 

With $n$ customer requests represented as pickup node (point) set $P = \{x_i\}_{i=1}^n$ and delivery node (point) set $D = {\{x_i\}_{i=n+1}^{2n}}$, the pickup node $x_i$ and the delivery node $x_{i+n}$ are bound as a pair with precedence relationship. With node 0 corresponding to depot, the complete node set is defined as $X=\{x_0\} \cup P \cup D$. Let $X'=X \cup \{x_{2n+1}\}=\{x_i \}_{i=0}^{2n+1}$, where node $x_{2n+1}$ is the copy of depot. Each node $x_i \in R^2$ is defined as $\{ c_i\}$, and $c_i$ contains 2-dimensional coordinates of location of $x_i$. 
We assume that all items picked from $x_i$ will be totally delivered to $x_{i+n}$ by a vehicle $v$ with infinity capacity (i.e., for the sake simplification).
With the above settings, the PDP describes a process that starting from the depot, a vehicle sequentially visits all pickup nodes and delivery nodes exactly once to perform the service, and finally returns to the depot, with the objective of minimizing the total travel time. Note that here PDP allows consecutive pickups, or deliveries, or mix of them as long as it satisfies the precedence constraint.


Let $D_{ij}$ be the Euclidean distance between node $x_i$ and node $x_j$, $f$ be the speed of vehicle $v$, $y_{ij}\in\{0,1\}$ be a binary variable to indicate whether the vehicle $v$ travels directly from node $x_i$ to node $x_j$, $B_i$ be the arrival time of node $x_i$, and $M$ be a large-enough number. Accordingly, the objective function of PDP is formulated as follows, 
\begin{equation}
     min\ \sum_{i \in X} \sum_{j \in X} \frac{D_{ij}}{f} y_{ij}.
     \label{eq:obj}
\end{equation}
Meanwhile, the PDP satisfies the following constraints,
\begin{align}
\sum_{j \in X} y_{ij} = 1,\qquad i \in X',\label{constraint 1}
\end{align}
\begin{align}
\sum_{i \in X} y_{ij} = 1,\qquad j \in X',\label{constraint 2}
\end{align}
\begin{align}
B_j \geq B_i + \frac{D_{ij}}{f} - M(1-y_{ij}), \qquad i\in X', j \in X',\label{constraint 3}
\end{align}
\begin{align}
B_{i+n} \geq B_i + \frac{D_{i,i+n}}{f}, \qquad i \in P, i+n \in D,\label{constraint 4}
\end{align}
\begin{align}
y_{ij} = \left\{ 0,1 \right\},\qquad i\in X', j \in X',\label{constraint 5}
\end{align}
\begin{align}
B_i \geq 0,\qquad i\in X'.\label{constraint 6}
\end{align}

The objective is to minimize the total travel time for the vehicle $v$. Constraint (\ref{constraint 1}) and (\ref{constraint 2}) ensure that each node including the depot ($x_0$) and its copy ($x_{2n+1}$) is visited exactly once. Constraint (\ref{constraint 3}) indicates the arrival time calculation for two adjacent nodes in the route. Constraint (\ref{constraint 4}) refers to the precedence constraint, and guarantees that the arrival time of a pickup node is earlier than that of its delivery node. Constraint (\ref{constraint 5}) defines the binary decision variable and constraint (\ref{constraint 6}) imposes the non-negativity of the arrival time. Note that we focus on a single vehicle in this paper, based on which it could be easily extended to multiple vehicles.

\section{Methodology}
\label{sec:method}
In this section, we first reformulate the PDP as the form of reinforcement learning (RL), then a policy network based on the encoder-decoder structure is designed to learn node selection for solution construction, which is empowered by a heterogeneous attention mechanism. 

\subsection{Reformulation as RL Form}
\label{sec:method_mdp}
For solving PDP, the route construction process could be essentially deemed as a sequence of decision making, which can be naturally formulated as the form of RL and solved accordingly. We model such route construction process as a Markov Decision Process (MDP), where an example of the MDP is presented in Fig. \ref{fig:MDP}. In specific, the elements of the MDP, i.e., the state space, the action space, the transition rule and the reward function are defined as follows.

\textbf{State.} The state $s_t = (L_t)$ represents the partial solution constructed at time step $t$, where $L_t$ contains all visited nodes until step $t$, and $L_0$ refers to the depot. 

\textbf{Action.} During the route construction process, the action $a_t$ is represented as $(x_j)$, i.e., selecting node $x_j$ at step $t$. 

\textbf{Transition.} The next state $s_{t+1} = (L_{t+1}) = {(L_t; \{x_j\})}$ is originated from $s_t$ by selecting a node at step $t$, where ``$;$" means concatenating the partial solution at last step with the newly selected node. 

\textbf{Reward.} To minimize the total travel time of routes, we define the reward as the negative of the objective value, which is calculated by accumulating the negative value of travel time of all steps. Consequently, the reward is represented as $R = \sum_{t=1}^T \boldsymbol{r}_t$, where $\boldsymbol{r}_t$ is the negative value of the incremental travel time at step $t$. Suppose that node $x_i$ and $x_j$ are selected at step $t$ and $t+1$, respectively, then $\boldsymbol{r}_{t+1}$ is expressed as a $m$-vector as follows,
\begin{equation}
\hspace{-0,1cm} \boldsymbol{r}_{t+1} =\!r(s_{t+1}, a_{t+1})\!=\! r((L_{t+1}), (x_j)) = - \frac{D_{ij}}{f},
\label{eq:reward}
\end{equation}
where $D_{ij}/f$ is the time for traveling from node $x_i$ to $x_j$.

\textbf{Policy.} The stochastic policy $p_{\theta}$ automatically selects a node at each time step under the precedence constraint. This process is repeated iteratively until all pickup-delivery services are completed. The final outcome engendered by performing the policy is a permutation of all nodes, which prescribes the order of each node for the vehicle to visit, i.e., $\pi  = \{ {\pi}_0,  {\pi}_1,...,  {\pi}_T\}$. Based on the chain rule, the probability of an output solution is factorized as follows,
\begin{equation}
   P(\pi|X) = \prod_{t=0}^{T-1} p_{\theta}(\pi_t | X, \pi_{1:t-1}),
\end{equation}
where $X$ is the input of an problem instance. And the decision making about the node selection will be performed based on the learnt $p_{\theta}$.

\begin{figure}
\centering 
 	\setlength{\belowcaptionskip}{-0.5cm}
	\includegraphics[width=0.48\textwidth, trim={3mm 1mm 2mm 1mm},clip]{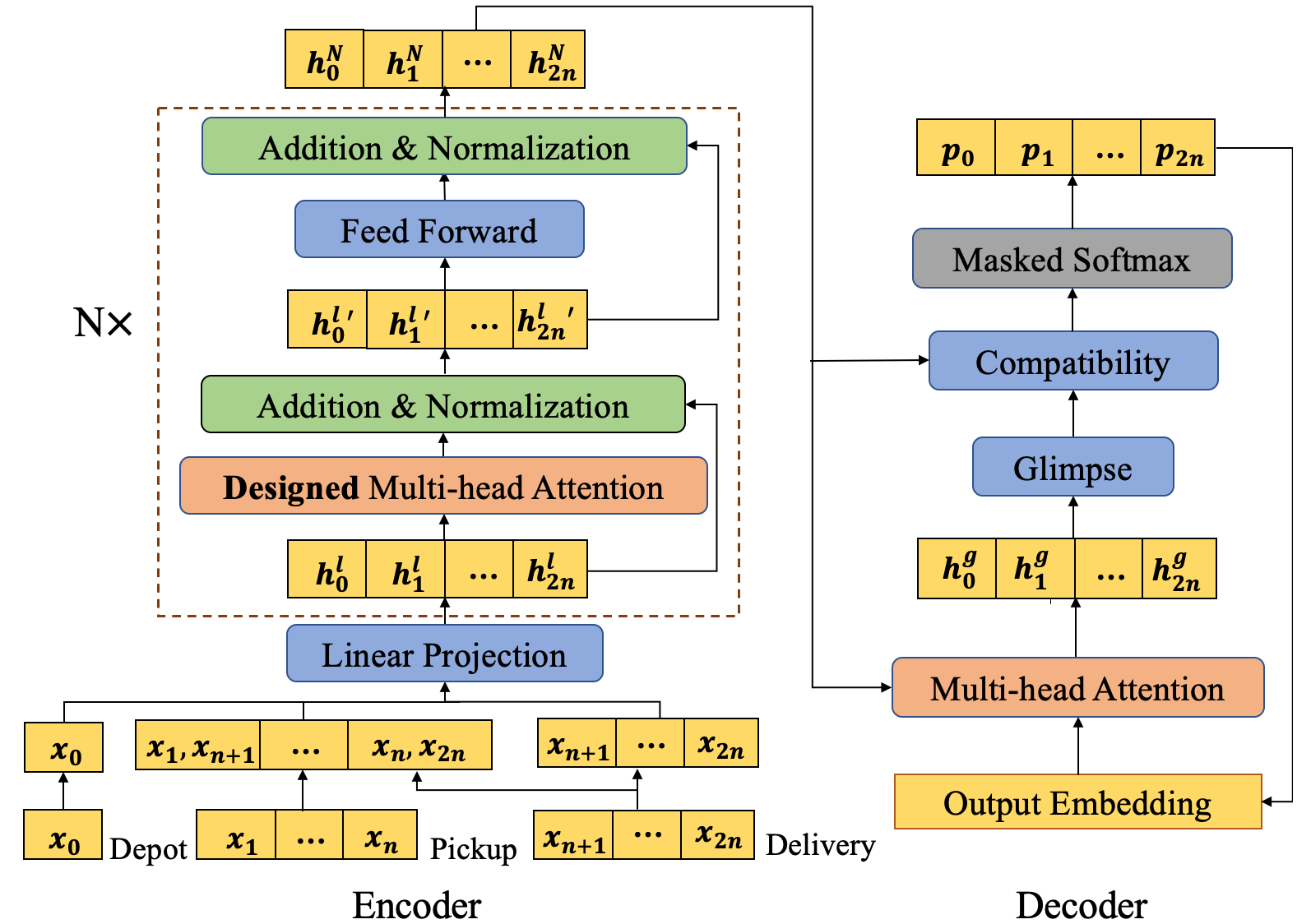} 
	\caption{Policy Network (left: encoder; right: decoder).} 
	\label{fig:DL_model} 
\end{figure}

\subsection{Policy Network based on Heterogeneous Attentions}
\label{sec:policynet}
The action space of a routing problem is discrete and grows exponentially as problem scales up, where the policy-based reinforcement method consisting an actor network and a critic network is often adopted to solve such problem. At each time step, the actor network generates a probability vector over all actions given the current state and then selects an action accordingly, which is iteratively repeated until the terminal condition. The reward of the actor network is calculated by summing up the accumulative reward at each step for the whole process. The critic network, as a baseline of the actor network, calculates the baseline reward only depending on the initial state to reduce variance. After receiving the reward of the actor network and the baseline reward of the critic network, the policy gradient method is adopted to update the parameters of two networks accordingly, where the actor network is trained towards finding solutions with higher quality.

To learn the policy $p_{\theta}$, we design a policy network based on an encoder-decoder structure as depicted in Fig.~\ref{fig:DL_model}, which is integrated with a heterogeneous attention scheme. Given the properties of PDP, the heterogeneous attentions are supposed to learn the relationship among the nodes of different roles, so that the precedence constraint could be intrinsically captured. Consequently, the parameters of this deep architecture also refer to the ones in $p_{\theta}$.

\subsubsection{Encoder} 
\label{sec:encoder}

\begin{figure*}
 \centering
	\includegraphics[width=1.0\textwidth, height=78mm, trim={0mm 1mm 1mm 1mm},clip]{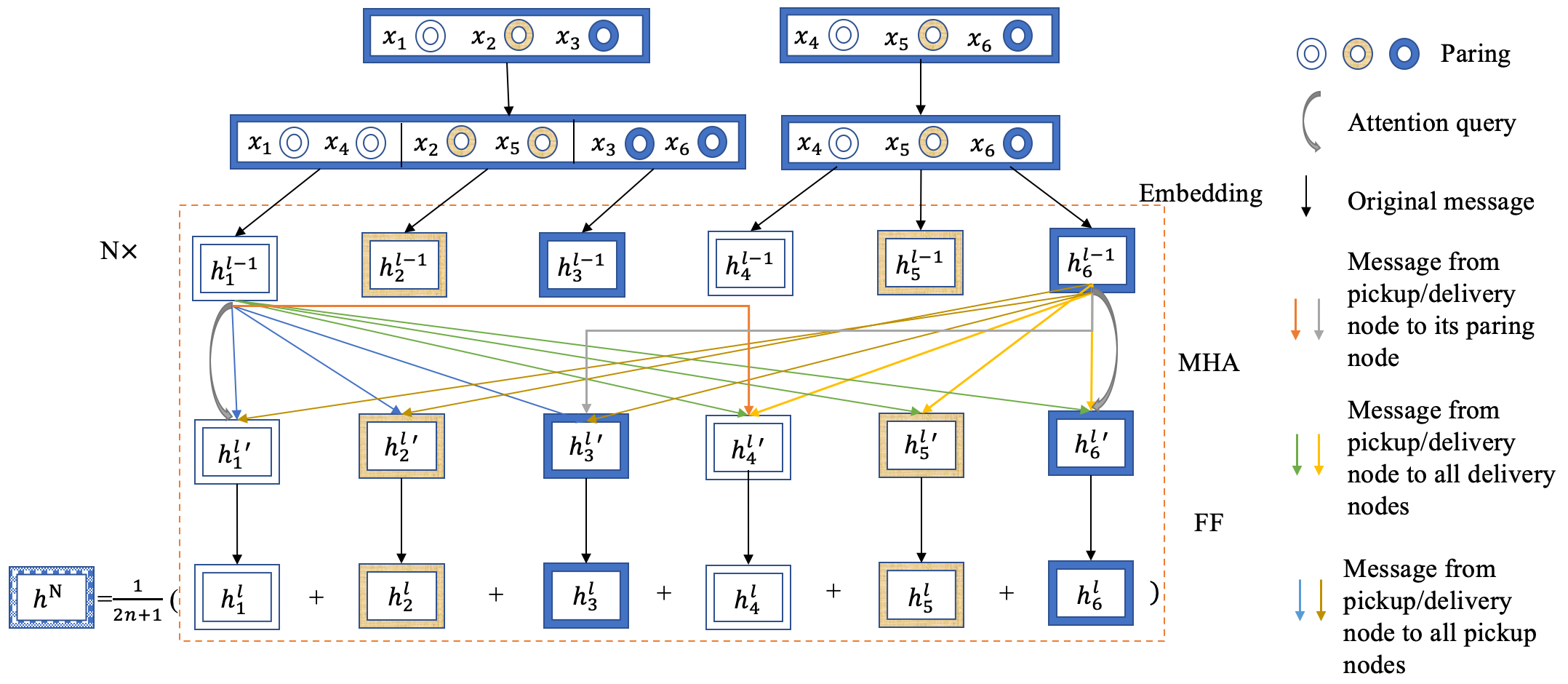} 
	\captionsetup{justification=justified}
	\caption{Illustration of the encoder. Taking pickup node $x_1$ as an example, it is concatenated with its delivery node $x_4$ to strengthen the pairing relation. After embedded into 128 dimensions, three types of attentions are calculated between node $x_1$ and other nodes of heterogeneous roles, i.e., the paired delivery node $x_4$, all pickup nodes ($x_1, x_2, x_3$) and all delivery nodes ($x_4, x_5, x_6$), respectively. The new attentions are supposed to capture the precedence relation and heterogeneous roles of nodes.
	} 
	\label{fig:Encoder} 
\end{figure*}

Regarding the encoder, we adopt a structure similar to the one in~\cite{kool2018attention}, with which we integrate a designed heterogeneous attention mechanism. Before elaborating the attentions, we first enhance the representations of inputs to the embedding by taking into account the pairing relation between the pickup node and delivery node. To this end, we concatenate the pickup nodes with the corresponding delivery nodes to strengthen the pairing relation, i.e., $x_i = (x_i; x_{i+n}), \forall \ i \in P$, where ``$;$" refers to concatenation of two vectors. The enhanced inputs are then linearly projected to initial node embeddings $h^{0}$ with dimension $d_h=128$. Afterwards the node embeddings go through $N$ attention layers, each of which encompasses a multi-head attention sublayer and a feed-forward (FF) sublayer.

Regarding the multi-head attention sublayer, we design heterogeneous attentions to better capture the precedence relation, and the details of the architecture are presented in Fig.~\ref{fig:Encoder}, where we leave out the depot for simplification. Let $h_i^{l-1}, i \in X$ denote the node embedding of attention layer $l-1 \ (l \in \{ 1,...,N \})$, where $i$ is the node index. Let $d_k$, $d_v$ be the \emph{query/key} dimension and $d_v$ be the \emph{value} dimension, where $d_k=d_v=\frac{d_h}{M}$ and $M=8$ is the number of heads. 

Given the input sequence $X$, the self-attention mechanism learns the relations between arbitrary two elements of the sequence to compute a representation of this sequence for better feature extraction~\cite{vaswani2017attention}. To capture the relations, three vectors, i.e., \emph{query, key} and \emph{value} are created based on the input $X$, which are expressed as follows,
\begin{equation}
   \! Q_i \!=\! W^Q h_i^{l-1},\  K_i \!=\! W^K h_i^{l-1},\  V_i \!=\! W^V h_i^{l-1}, \ i \in X,
\end{equation}
where $W^Q, W^K \in R^{d_h \times d_k}$ and $W^V \in R^{d_h \times d_v}$ are trainable parameter matrices, and $Q$, $K$ and $V$ refer to \emph{query}, \emph{key} and \emph{value}, respectively.
The scaled dot product is selected as a compatibility function of the \emph{query} and the \emph{key} to measure the importance between arbitrary two nodes, then the compatibility is further processed by a softmax function to compute the weights of all nodes. Following this, the original self-attention for solving the typical VRP in~\cite{kool2018attention} is calculated as follows,
\begin{equation}
 \hspace{-0,1cm} a_{ij} = softmax(\frac{Q_i^{T} K_j}{\sqrt{d_k}}), \  i,j \in X,
\end{equation}
where higher attention value $a_{ij}$ means that the node $x_i$ depends more on the node $x_j$.

However, different from the typical VRP, the nodes in PDP may have heterogeneous roles, and thus impose more complex relations. For instance, the roles of pickup node and delivery node are different; the pickup node must precede the pairing delivery node; multiple pickup (or delivery) nodes are allowed to be visited consecutively, etc. Therefore, homogeneously treating those nodes with an attention may limit the quality of the solutions. To alleviate this issue, we design six types of attentions in addition to the original one. Among them, three types of attentions are constructed for each pickup node to learn the relations to its pairing delivery node, all pickup nodes and all delivery nodes, respectively. And another three types of attentions are constructed for each delivery node in a similar way. 

The details of the heterogeneous attention scheme are as follows. Suppose that $h_i^P\ $ is the embedding of the $i$th pickup node, and $h_j^D\ $ is the embedding of the $j$th delivery node, then the node embedding of layer $l-1$ can be expressed as $h^{l-1} = Concat(h_0,h_i^P, h_j^D), \forall \ i \in P, \forall \ j\in D$, where we omit $l-1$ for readability. Accordingly, the core of attentions between each pairing nodes is calculated as follows,
\begin{equation}
\begin{split}
    & \!\!Q_i^{pd} \!= \!W^{Q_{pd}} h_i^P, \ K_i^{pd}\! =\! W^K h_{i+n}^D, \ V_i^{pd} \!= W^V\! h_{i+n}^D, \\
    & \!\!Q_i^{dp} \!= \!W^{Q_{dp}} h_i^D, \ K_i^{dp}\! =\! W^K h_{i-n}^P, \ V_i^{dp} \!= W^V\! h_{i-n}^P.
\end{split}
\end{equation}

Similarly, the core of attentions from each pickup node to arbitrary pickup nodes and arbitrary delivery nodes is calculated as follows,
\begin{equation}
\begin{split}
    &\!\!Q_i^{pP} \!=\! W^{Q_{pP}} h_i^P, \ K_i^{pP} \!=\! W^K h_i^P, \ V_i^{pP} \!=\! W^V h_i^P, \\
    & \!\!Q_i^{pD} \!=\! W^{Q_{pD}} h_i^P, \ K_i^{pD} \!=\! W^K h_i^D, \ V_i^{pD} \!=\! W^V h_i^D,
\end{split}
\end{equation}
and the core of attentions from each delivery node to arbitrary pickup nodes and arbitrary delivery nodes is calculated as follows,
\begin{equation}
\begin{split}
    &\!\!Q_i^{dP} \!=\! W^{Q_{dP}} h_i^D, \ K_i^{dP} \!=\! W^K h_i^P, \ V_i^{dP} \!=\! W^V h_i^P, \\
    & \!\!Q_i^{dD} \!=\! W^{Q_{dD}} h_i^D, \ K_i^{dD} \!=\! W^K h_i^D, \ V_i^{dD} \!=\! W^V h_i^D,
\end{split}
\end{equation}
where all parameter matrices are trainable and have same sizes with the original one. Note that we share parameter matrices of all \emph{keys} and \emph{values} for the seven types of attentions to speed up the training, while keep all parameter matrices of \emph{queries} independently reserved to learn problem properties from different perspectives. 

Consequently, the immediate attentions of different types could be achieved based on above cores. Regarding the two types of paired nodes, the attentions are calculated as follows,
\begin{equation}
\begin{split}
    &a_{i,i+n}^{pd} = softmax(\frac{Q_i^{pd} * K_{i+n}^{pd}}{\sqrt{d_k}}), \ i \in P, \\
    & a_{i,i-n}^{dp} = softmax(\frac{Q_i^{dp} * K_{i-n}^{dp}}{\sqrt{d_k}}),\ i \in D,
\end{split}
\end{equation}
where "*" means the element-wise product. Regarding the remaining four types, the immediate attentions are calculated as follows,
\begin{equation}
 \hspace{-0,1cm} a_{ij}^y = softmax(\frac{Q_i^{yT} K_j^y}{\sqrt{d_k}}),\ y \in \{pP, pD, dP, dD \}.
\end{equation}
Then, the multi-head vector concatenates different information from $M$ heads as follows,
\begin{equation}
\text{ MultiHead} (Q_i^y, K_j^y,V_j^y)\!=\! Concat(h_i^1, ..., h_i^M) W^O,
\end{equation}
where $W^O \in R^{d_h \times d_h}$ is the trainable parameter matrices, and the single head vector $h_i$ is calculated by adding up the heads from all types of attentions as follows,    
\begin{equation}
\begin{split}
    \hspace{-0,1cm} h_i^m
    & = a_{ij} V_j + a_{ij}^{pd} * V_j^{pd}+ a_{ij}^{dp} * V_j^{dp} + \sum_y \sum_j a_{ij}^y V_j^y,\\
    & y \in \{pP, pD, dP, dD\}, \ m \in \{ 1,...,M \}.
\end{split}
\end{equation}
Note that to enable the policy network to perceive and learn the pairing relationships and the precedence constraint, the heads are added to different parts of node embeddings according to the types of attentions. E.g., the heads with attentions from pickup nodes to other roles of nodes only contribute to the pickup node embeddings, where the heads added to the delivery node embeddings are all zero.

Finally, the attention mechanism works as follows,
\begin{equation}
 {h_i^l}' = BN^l(h_i^{l-1} + \text{MultiHead}_i^l (Q_i^y, K_j^y,V_j^y)),
\end{equation}
\begin{equation}
 h_i^l = BN^l({h_i^l}' + FF^l({h_i^l}'),
\end{equation}
where $h_i^l$ is the node embeddings at the $l$th layer, and parameters are independently reserved for different layers. Each multi-head attention layer and feed-forward layer consists of a skip-connection~\cite{he2016deep} and a batch normalization (BN) layer~\cite{ioffe2015batch}. 

Subsequently, the graph embedding of inputs $h^N$ is computed as the mean of node embeddings at final layer, i.e., $\bar{h}^N = \frac{1}{2n+1} \sum_{i=0}^{2n} h_i^N$. Both the node embeddings $h_i^N$ and graph embedding $\bar{h}^N$ are taken as the input of the decoder.

\subsubsection{Decoder}
Given the graph embedding and node embeddings from the encoder, the decoder will generate a probability vector for selecting a node at each decoding step. 

To achieve this, a \emph{context} $h^c$ (output embedding) is always needed at the beginning, which only consists of graph embedding and last node embedding at time step $t$ as follows, 
\begin{equation}
 h^c = Concat(\bar{h}^N, h_{\pi_{t-1}}^N).
\end{equation}
At the first time step, the node embedding is usually replaced with trainable parameters. Similar to~\cite{vinyals2015order}, the \emph{glimpse} $h^g$ used to aggregate the contributions from different parts of node information is expressed as follows,
\begin{equation}
 h^g = \text{MultiHead}(W_g^Q h^c, W_g^K h^N, W_g^V h^N),
\end{equation} 
where $W_g^Q, W_g^K \in R^{d_h \times d_k}, W_g^V \in R^{d_h \times d_v}$ are trainable parameter matrices. 
Given that $q=W^Q h^g$ and $\ k_i=W^K h_i^N$, the \emph{compatibility} with all nodes at step $t$ is calculated as follows, 
\begin{equation}
 \hat{h^t}=C \cdot tanh(h^t),
\end{equation} 
where
\begin{equation}
h_i^t\!=\!\left\{
\begin{array}{lr}
\frac{q^T k_i}{\sqrt{d_k}},\ & if\  i \notin \pi_{t^{\prime}},\  {\forall}\  t^{\prime}\! < t, \\
-{\infty},\! & {\rm otherwise},
\end{array}
\right.
\end{equation}
and $C$ is set to 10 to clip the result for better exploration. Meanwhile, all invalid nodes are dynamically masked at each step to guarantee feasibility. In particular, upon departure from the depot, all delivery nodes would be masked. After visiting the first pickup node, the pairing delivery node would be unmasked, and the visited nodes, including the depot, would be masked. The masked nodes are allowed to be unmasked only when feasibility is satisfied. Finally, we compute the probability vector using the softmax function as follows,
\begin{equation}
 p(\pi_t|X, L_{t-1})=softmax(\hat{h^t}),
\label{eq:prob}
\end{equation} 
where element $p_i^t$ represents the probability of selecting node $x_i$ at step $t$. This process iteratively continues until all nodes are visited and the vehicle returns to the depot. 
Regarding the decoding strategy, we could choose the node with the maximum probability at each step in a greedy manner, and we could also sample multiple solutions and retrieve the best one, which will be investigated in the experiments.

\IncMargin{0.5em}
\begin{algorithm} [!t]
\small
\SetKwData{Left}{left}\SetKwData{This}{this}\SetKwData{Up}{up}
\SetKwFunction{Union}{Union}\SetKwFunction{FindCompress}{FindCompress}
\SetKwInOut{Input}{input}\SetKwInOut{Output}{output}
\Input{number of episodes $I$; 
    batched size $B$; 
    step limits for route constructions $T$;\\
    actor network $p_\theta$ with parameters $ \theta$;\\     
    critic network $v_\phi$ with parameters $\phi$; \\
    }
\ForEach{$epoch=1,2,...,I$ }
{
 Generate $B$ problem instances randomly;\\
 Reset gradients $d_{\theta} \leftarrow 0$;\\
 \ForEach{$b=1,2,...,B$}
 {
     Initiate state $s_t^b=(L_0^b)$;\\
     \While{$t<T$}
     {
      $a_t^b \sim p_{\theta}(\cdot | s_t^b)$;\\
      Pick an action $a_t^b$ in sampling strategy using the policy network $p_{\theta}$;\\
      Receive reward $r_t^b$ and transit to next state;\\
      $t=t+1$;\\
     }
  $R^b = \sum\limits_{t=1}^T r_t^b$;
 
 Receive baseline reward $v_{\phi}(X^b)$ using GreedyRollout policy $v_{\phi}$; \\
 }
 $d_{\theta} \leftarrow \frac{1}{B} \sum\limits_{b=1}^B (R^b - v_{\phi}(X^b)) \nabla_{\theta} log\  p_{\theta}(\pi^b|X^b)$;\\
 Update $\theta$ using $d_{\theta}$;\\

\If{OneSidedPairedTTest($p_{\theta}, v_{\phi})\  \textless \alpha$} 
{
 Replace $v_{\phi}$ with $p_{\theta}$;
}}
\caption{Reinforcement Learning Algorithm}\label{Algorithm:algorithm}
\end{algorithm}

\subsection{Training Algorithm}

The training of the proposed policy for solving PDP is summarized in Algorithm~\ref{Algorithm:algorithm}, where we adopt the reinforcement learning method with roll-out baseline in~\cite{kool2018attention}. The policy gradient method is characterized by two networks: 1) actor network, i.e., the $p_{\theta}$ mentioned above, governs the actions of node selection by generating a vector of probabilities over those actions, and sampling according to the probabilities to better explore the action space; 2) self-critic network $v_{\phi}$, i.e., a roll-out baseline with similar structure as the actor network, calculates the reward given the initial state by selecting the node with maximum probability to eliminate variance. After receiving the reward of the actor network $R$ and the baseline reward of the critic network $v_{\phi}(X)$, the reinforcement learning algorithm updates the parameters of two networks accordingly using the policy gradient method. In specific, at each episode, we construct a route for each instance and calculate the reward with respect to this solution in line 12, and the parameters of actor network are updated in line 16. Moreover, the expected reward of critic network $v_{\phi}(X^b)$ for instance $b$ is obtained from a greedy roll-out of the policy in line 13. Furthermore, the parameters of critic network are replaced with that of actor network when the performance of the latter is significantly superior according to a paired t-test on several fixed number of instances in line 18~\cite{kool2018attention}. By updating the two networks, the policy $p_{\theta}$ is trained towards finding solutions of higher quality.


\section{Computational Experimentation and Analysis}
\label{sec:experiments}

In this section, we conduct experiments to verify the performance of the proposed DRL model for solving PDP. Specifically, with the customers divided into paired pickup and delivery nodes, a vehicle starts at depot, and visits all customers exactly once with pairing relations and precedence constraint, i.e., the pickup node must be visited before its delivery node. The objective is to minimize the total travel time. Note that, PDP is an NP-hard problem, and the computation complexity grows exponentially as problem scales up. 

\begin{figure}[t] 
\centering 
 	\setlength{\belowcaptionskip}{-0.4cm}
	\includegraphics[width=0.45\textwidth,trim=0 15 0 30]{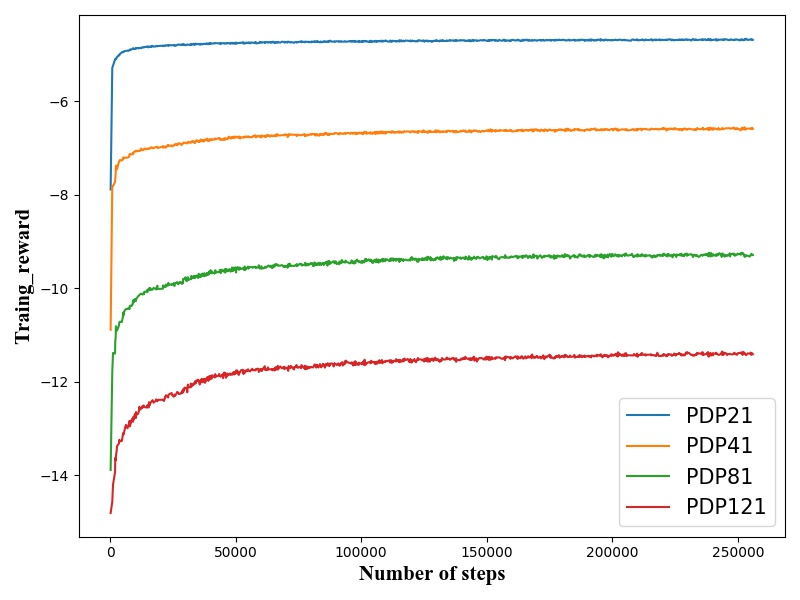} 
	\caption{The Reward Curve.}
	\label{fig:reward} 
\end{figure}

\subsection{Experimentation Settings} 
Following the settings in existing works~\cite{bello2017neural,nazari2018reinforcement,kool2018attention,chen2019learning}, the locations of the depot and customer (pickup and delivery pair)  nodes are randomly and independently generated using a 2-dimensional uniform distribution in the range of 0 and 1, where the distance between two nodes are calculated based on Euclidean space. For expression simplification, the vehicle speed $f$ in Eq.~(\ref{eq:obj}) and Eq.~(\ref{eq:reward}) is fixed to 1, however, our method is capable of handling varying speeds. Regarding the scale, the problem sizes are set to 21, 41, 81, and 121 (including one depot and all nodes of pickup and delivery), and termed as PDP21, PDP41, PDP81, PDP121, respectively.

Regarding training, all instances are generated on the fly. We run 800 epochs, and in each epoch, 2500 batches with 512 instances are processed. For each instance, the 2-dimension locations of nodes are first linearly projected to a 128-dimension vector, then processed by a 3-layers heterogeneous attention mechanism before fed into the decoder that shares the same hidden layer dimension with the encoder. Moreover, we adopt the Adam Optimizer to train the policy network with constant learning rate $10^{-4}$. Each epoch takes 4.37 mins for PDP21, 27.48 mins for PDP41 with single GPU (2080Ti) 35.47 mins for PDP81 with two GPUs and 44.11 mins for PDP121 with three GPUs, respectively.
Regarding testing, instances are fixed for our method and all baselines, where 10000 instances are generated for each problem size, using the same distribution with the training one. 
\begin{table}
\caption{Diverse DRL models for PDP.}
\centering
\resizebox{0.48\textwidth}{!}{
\begin{tabular}{c|c c c|c c c} 
\toprule
& 
\multicolumn{3}{c|}{PDP81} & 
\multicolumn{3}{c}{PDP121} \\
Method & Obj.  & Gap & Time & Obj.   & Gap  & Time    \\
\midrule
4 attns-no-sharing & 9.009	& 0.54\% & 1.72s &11.024  & 0.57\% & 3.64s\\
4 attns-sharing  & 8.987 & 0.29\%  & 1.72s &11.371  & 3.73\% & 3.69s  \\
7 attns-no-sharing & 9.001	& 0.45\% & 1.78s &10.965  & 0.03\% & 3.78s \\
7 attns-sharing  & 8.961	& 0\% & 1.94s  &10.962  & 0\%  & 3.83s\\
\bottomrule
\end{tabular}}
\label{tab:diversity}
\end{table}

\begin{table*}
\caption{DRL Method VS Baselines for PDP.}
\centering
\resizebox{\textwidth}{!}{
\begin{tabular}{c|c c c|c c c|c c c|c c c} 
\toprule
& 
\multicolumn{3}{c|}{PDP21}   &
\multicolumn{3}{c|}{PDP41} & 
\multicolumn{3}{c|}{PDP81} & 
\multicolumn{3}{c}{PDP121} \\
Method & Obj.   & Gap   & Time   & Obj.   & Gap   & Time  & Obj.   & Gap   & Time & Obj.   & Gap   & Time  \\
\midrule
CPLEX  & \textbf{4.560} & \textbf{0\%}  & 433.72s & - & -	& -	& -	& - & - &-  & -	& - \\
SA & 4.602 & 0.92\%  & 48.14s & 6.409 & 1.50\%	& 155.44s	& 9.098	& 1.53\% & 305.58s &11.298  & 3.07\%	& 497.70s \\
VNS &  4.608 & 1.05\%  & 90.78s  & 6.592 & 4.40\%	& 139.25s	& 9.732	& 8.60\% & 583.90s &12.307  & 12.27\%	& 1035.83s \\
OR-Tools  & 4.704 & 3.16\%  & 0.29s & 6.586 & 4.31\%	& 7.45s	& 9.188	& 2.53\% & 233.72s &11.173  & 1.92\%	& 1774.58s \\
AM(Greedy)  & 5.021 & 10.11\%  & \textbf{0.09s} & 7.180 & 13.72\%	& \textbf{0.15s}	& 10.042	& 12.06\% & \textbf{0.37s} &12.334  & 12.52\%	& \textbf{0.52s} \\
AM(Sample1280)  & 4.731 & 3.75\%  & 0.26s & 6.739 & 6.73\%	& 0.43s	& 9.469	& 5.67\% & 0.91s &11.674  & 6.50\%	& 1.78s \\
AM(Sample12800) & 4.691 & 2.87\%  & 0.64s & 6.673 & 5.69\%  & 1.00s & 9.382 & 4.70\%	 & 1.83s &11.565  & 5.50\%	& 3.51s \\
DRL(Greedy)	& 4.668 & 2.37\%  & 0.11s  & 6.536 & 3.52\%	& 0.19s	& 9.229	& 2.99\% & 0.41s &11.322 & 3.50\%	& 0.70s \\
DRL(Sample1280)  & 4.593 & 0.72\%  & 0.29s  & 6.338 & 0.38\%	& 0.49s	& 8.996	& 0.39\% & 0.98s &11.025  & 0.84\%	& 1.90s \\
DRL(Sample12800) & 4.585 & 0.55\%  & 0.84s  & \textbf{6.314} & \textbf{0\%}	& 1.12s	& \textbf{8.961}	& \textbf{0\%} & 1.94s &\textbf{10.962}  & \textbf{0\%}	& 3.83s \\       
\bottomrule
\end{tabular}}
\label{tab:comparison}
\end{table*}

\subsection{Diversity of the Heterogeneous Attentions}
With respect to the proposed DRL model, we apply two versions of decoding strategy during testing: 1) \emph{Greedy}, always selects the node with maximum probability at each decoding step under the precedence constraint; 2) \emph{Sampling}, engenders $\mathcal{N}$ solutions for each instance by sampling the probability distribution in Eq.~(\ref{eq:prob}), and retrieves the best one, where $\mathcal{N}$ is set to 1280 and 12800, and termed as DRL(\emph{Sample1280}) and DRL(\emph{Sample12800}), respectively. Before comparing with others, we first depict the training curves of our DRL method for all problem sizes in Fig.~\ref{fig:reward}. From the reward curves we observe that the rewards increase very fast for all cases until converged. As the number of customers increases, the rewards become smaller and converge relatively slower, which is reasonable since reward is inversely proportional to the route length and the computation complexity grows when the problem scales up.

To verify the impacts of different attention mechanisms on the performance, we evaluate different combinations of attentions. Apart from the seven types of attentions in our method, we also consider a mechanism with four types of attentions. Regarding the latter, besides the original attention, it only integrates the attentions from each ego pickup node to nodes of other three roles, i.e., the ego delivery node, all pickup nodes and delivery nodes. We term them as \emph{7 attns-sharing} (ours) and \emph{4 attns-sharing}, respectively. For the two mechanisms, we also consider their variants by disabling the parameter sharing for \emph{keys} and \emph{values}, which are termed as \emph{7 attns-no-sharing} and \emph{4 attns-no-sharing}, respectively. We concentrate on Sample12800 for PDP81 and PDP121, and the results are recorded in Table~\ref{tab:diversity}. We observe that the policy with 7 attns-sharing (ours) achieves the smallest objective value with only slightly longer computation time, which well justified the rationale for the design of our heterogeneous attentions. Therefore, we adopt \emph{7 attns-sharing} in our method.

\subsection{Comparison Analysis}
\label{sec:comparison}
The proposed DRL model is compared with five baselines: 1) CPLEX~\cite{cplex}, a state-of-the-art exact solver for combinatorial optimization problems; 2) OR-Tools~\cite{ortools}, a constraint-solver based method that widely used for combinatorial optimization problems; 3) Simulated Annealing (SA)~\cite{yu2016solving}, a metaheuristic method with guided neighborhood search for solving location routing problem with simultaneous pickup and delivery; 4) Variable Neighborhood Search (VNS)~\cite{pinto2017variable}, a heuristic method based on  different neighborhood structures for solving PDP with loading constraints; 5) the DRL based attention model (AM)~\cite{kool2018attention}, learning a policy of node selection in a constructive manner for TSP and CVRP. The time limit of CPLEX for solving PDP20 is set to 500 seconds for reasonable computation time.
We adapt the pairing and precedence constraints of PDP to all baselines so that they share the same problem characteristics with ours.

We record the performance of our DRL method and baselines for all problem sizes in Table \ref{tab:comparison}, where the objective value, optimality gap, and computation time of an instance are adopted for evaluation. Note that, according to our experiments, the exact solver CPLEX is only available for PDP21, and prohibitively time-consuming for larger sizes. Therefore, the gap is defined by comparing with CPLEX for PDP21, and with the best one among all methods for other sizes. From Table \ref{tab:comparison}, we observe that our DRL method with Sample12800 outperforms all conventional heuristic methods and AM model for all cases. Although the computation time of our method is slightly longer than that of AM, it is significantly shorter than that of all conventional baselines. Regarding three variants of our DRL method and AM, both Sample1280 and Sample12800 produce smaller objective values and gaps than Greedy, which demonstrates the effectiveness of sampling strategy in improving the solution quality, despite of slightly longer computation time. Among all baselines, CPLEX can engender optimal solutions for PDP21, but suffers from the longest computation time for solving a single instance, i.e., 433.72 seconds. SA achieves better performance in terms of objective values and gaps than other heuristic methods for most cases except for PDP121, where OR-Tools achieves smaller objective value but longer computation time. Furthermore, AM(Greedy) achieves shortest computation time and highest objective values. Moreover, given that Sample12800 is superior to Sample1280, and Sample1280 superior to Greedy, our Greedy alone outperforms all variants of AM regarding objective values and gaps, which verifies that the proposed heterogeneous attention mechanism is capable of empowering the policy network to intrinsically capture and learn the precedence relations. Another phenomenon is that, even our sole Sample1280 outperforms all conventional heuristic methods and AM in all aspects. Besides, as the problem scales up, the objective value becomes larger and computation time becomes longer for all methods, however, the computation time of DRL based methods almost increases linearly, with that of conventional methods growing exponentially. From Table \ref{tab:comparison}, our DRL method achieves the best overall performance in comparison with AM and all conventional methods.

\begin{table*}
\caption{Generalization on a Different Distribution (i.e.,Gaussian Distribution).}
\centering
\resizebox{\textwidth}{!}{
\begin{tabular}{c|c c|c c|c c|c c|c c|c c} 
\toprule
&
\multicolumn{2}{c|}{PDP81(sdv=0.6)}   &
\multicolumn{2}{c|}{PDP121(sdv=0.6)} & \multicolumn{2}{c|}{PDP81(sdv=0.8)} & \multicolumn{2}{c|}{PDP121(sdv=0.8)} &
\multicolumn{2}{c|}{PDP81(sdv=1.0)}   &
\multicolumn{2}{c}{PDP121(sdv=1.0)}   \\
Method  & Obj.   & Time   & Obj.  & Time  & Obj.  & Time & Obj.   & Time & Obj.  & Time  & Obj. & Time   \\
\midrule
SA	& 8.916 & 342.05s &11.092	& 505.84s & 8.984 & 344.79s & 11.205	& 507.03s & 9.013	& 364.59s & 11.228	& 515.68s \\
OR-Tools & 9.027	 & 233.83s &11.048 	& 1740.23s & 9.103& 242.38s &11.119  & 1747.34s & 9.129 & 244.07 &11.139 	& 1753.76s \\
AM(Greedy) & 9.882	& 0.31s &12.188  & 0.55s & 9.954 & 0.31s &12.257 	& 0.57s & 9.987 & 0.32s &12.288  & 0.56s \\
AM(sample1280) & 9.291	& 0.83s &11.518  & 1.93s & 9.372 & 0.83s &11.592 	& 1.93s & 9.406 & 0.84s &11.622  & 1.94s \\
AM(sample12800) & 9.204	& 1.76s &11.405  & 3.83s & 9.283 & 1.77s &11.481 	& 3.85s & 9.318 & 1.78s &11.512  & 3.88s \\
DRL(Greedy) & 9.082	& 0.42s &11.141  & 0.70s & 9.145 & 0.42s &11.219 	& 0.69s & 9.173& 0.41s &11.253  & 0.68s \\
DRL(sample1280) & 8.844	& 0.89s &10.840	& 1.97s & 8.911	&  0.89s &10.919 	& 1.98s & 8.940 & 0.88s &10.953  	& 1.99s\\
DRL(sample12800) & 8.807& 1.94s &10.787  	& 3.95s & 8.876 & 1.93s &10.866 	& 3.98s & 8.904 &  1.94s &10.900  &  3.97s \\      
\bottomrule
\end{tabular}}
\label{tab:comparison_sdv}
\end{table*}

\begin{figure}[t] 
\centering 
 	\setlength{\belowcaptionskip}{-0.4cm}
	\includegraphics[width=0.45\textwidth,trim=15 20 5 20]{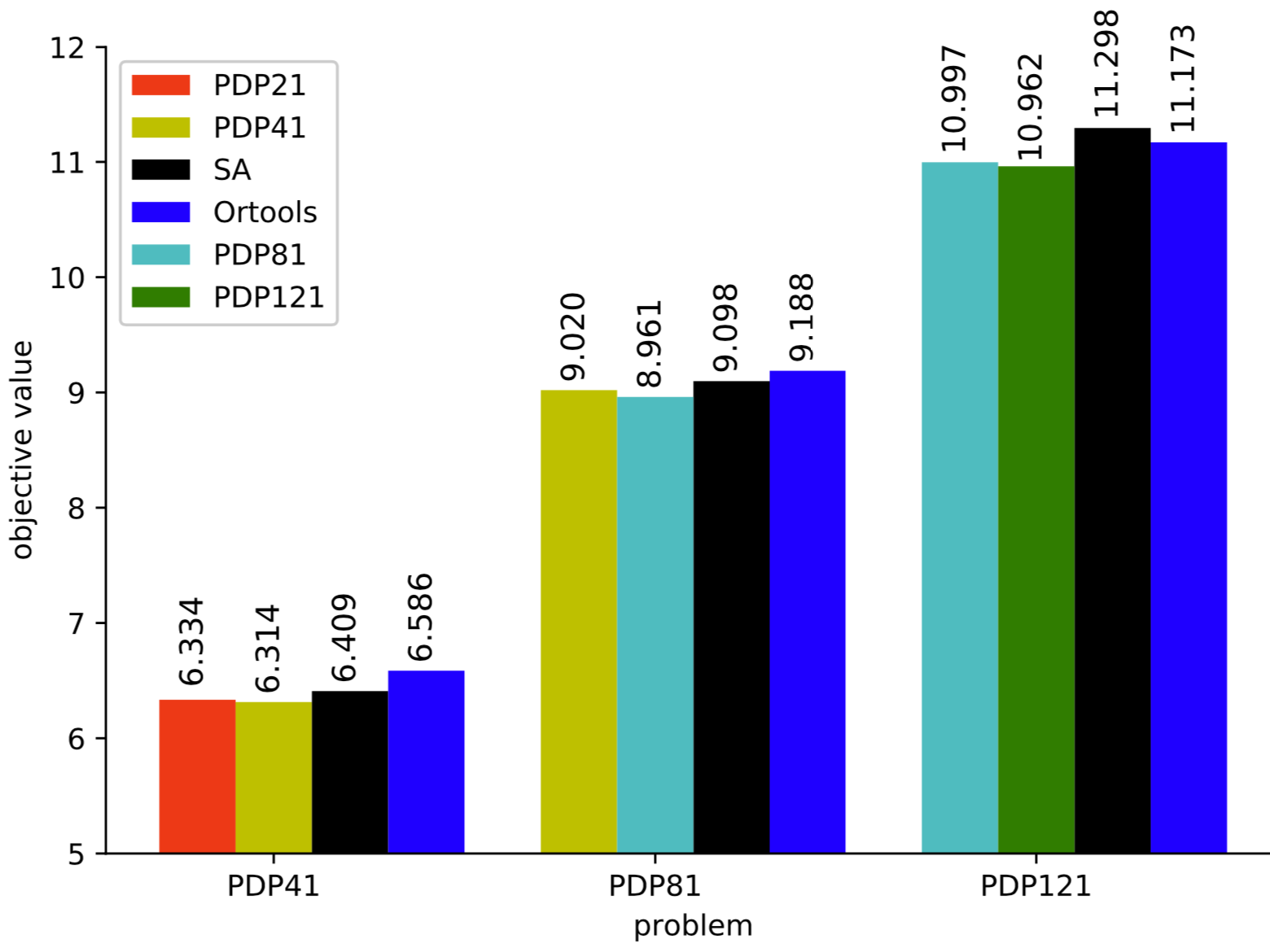} 
	\caption{Generalization on larger problem size.}
	\label{fig:generalization} 
\end{figure}

\subsection{Generalization Analysis}
To verify the generalization of our method, we continue to conduct two types of experiments: 1) apply the policy learnt for a problem size to a larger one, 2) apply the policy learnt for uniform distribution to others, i.e., Gaussian distribution. 

Regarding the former, we focus on sample12800, and apply the policy learnt for a problem size to its larger neighbour, e.g., apply sample12800 for PDP21 to PDP41. In Fig.~\ref{fig:generalization}, the horizontal coordinate refers to the problems to be solved, and the vertical one refers to the objective values of different policies or methods, where we consider SA and OR-Tools given their good performance in Table \ref{tab:comparison}. We can observe that for each problem size, the corresponding policy performs best, which is slightly superior to the one learnt for a different size. However, both of them outstrip the best conventional methods, from which we conclude that our DRL method has desirable capability of generalization on larger problem sizes.

Regarding the latter, we focus on PDP81 and PDP121 since they are the largest two, where we apply the policies learnt with uniform distribution to Gaussian distribution with different standard deviations. In Table \ref{tab:comparison_sdv}, three standard deviations are adopted, and the one with 1.0 is closer to the original uniform distribution. 
We observe that our DRL method with Sample12800 always outperforms all other methods in all cases in terms of objective value, and AM(Greedy) consumes shortest computation time. Specifically, both Sample1280 and Sample12800 achieves smaller objective value than conventional heuristics with significantly shorter computation time, and AM with comparable computation time. SA performs better in terms of objective value than OR-Tools for PDP81 but consumes longer computation time, which is the other way round for PDP121. Given the strong superiority to the conventional heuristics, we conclude that our method has satisfactory generalization on a new distribution.

\section{Conclusion and Future Work}
\label{sec:conclusion}
In this paper, we cope with a challenging variant of VRP, i.e., the pickup and delivery problem (PDP), which is characterized by a precedence constraint that a pickup node must precede the pairing delivery node. To solve this problem, we propose a deep reinforcement learning method integrating with a heterogeneous attention mechanism, which empowers the policy network to capture and learn the precedence relation and heterogeneous roles of different nodes. Experimental results show that our DRL method achieves the best overall performance compared with existing modern (i.e., AM) and conventional heuristic methods, respectively. Moreover, our method generalizes well to larger problem size and new distribution. Additionally, our method with diverse combinations on heterogeneous attention mechanisms outperform AM and all conventional heuristic methods.  

Exploring deep reinforcement learning to solve vehicle routing problem is a newly emerged direction, which is still at a preliminary stage despite of the good potential. The work in this paper is an early attempt on the learning based method for solving PDP, which delivers superior results but may perform inferior for other types of VRPs, such as VRP with time windows constraint and dynamic customer requests. In the future, we plan to investigate the following aspects, 1) multiple vehicles with capacity constraint; 2) time window constraint and dynamic customer requests (or traffic conditions); and 3) real data or other classical benchmark datasets for testing.




\bibliographystyle{ieeetr}
\bibliography{PDP}

\vspace{-10mm}
\begin{IEEEbiography}[{\includegraphics[width=1.0in, height=1.2in, clip,keepaspectratio]{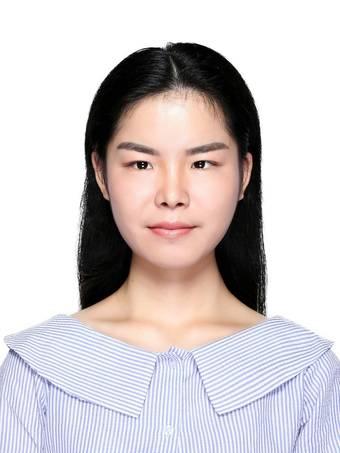}}]{Jingwen Li} received the bachelor’s degree in the field of computer science from University of Electronic Science and Technology of China, China, in 2018. She is currently pursuing the phd’s degree with the department of Industrial Systems Engineering and Management, National University of Singapore (NUS). Her research interests include deep reinforcement learning for combinatorial optimization problems, especially for vehicle routing problems.
\end{IEEEbiography}

\begin{IEEEbiography}[{\includegraphics[width=1.0in, height=1.2in, clip,keepaspectratio]{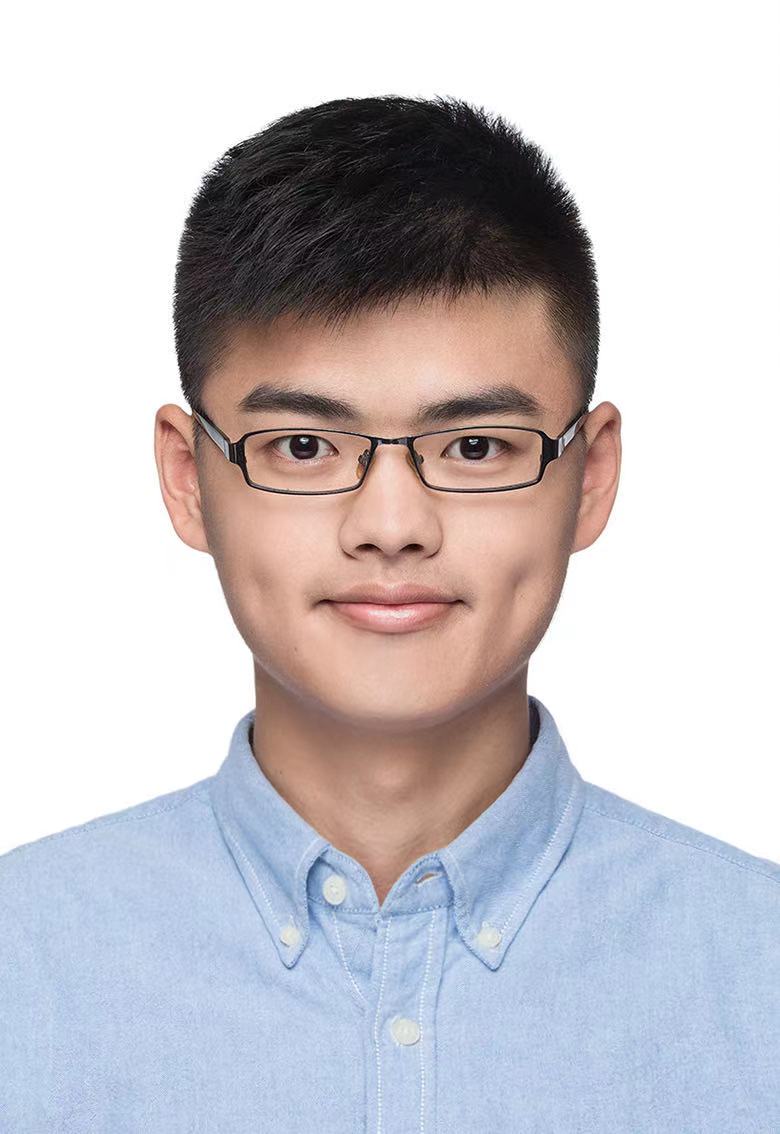}}]{Liang Xin} received the bachelor’s degree from Tongji University and the master's degree from Carnegie Mellon University. He is currently pursuing the phd’s degree with the School of Computer Science and Engineering, Nanyang Technological University, Singapore (NTU). His research interests include deep learning for combinatorial optimization problems.
\end{IEEEbiography}


\begin{IEEEbiography}[{\includegraphics[width=1.0in, height=1.2in, clip,keepaspectratio]{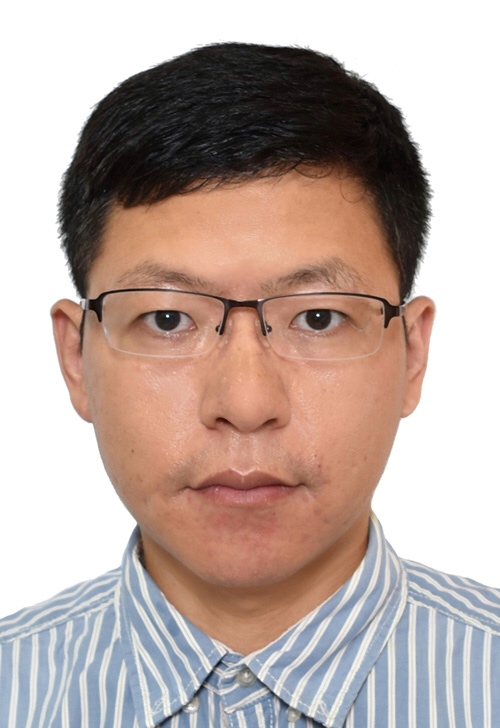}}]{Zhiguang Cao} received the Ph.D degree from Interdisciplinary Graduate School, Nanyang Technological University, Singapore, 2017. He received the B.Eng. degree in Automation from Guangdong University of Technology, Guangzhou, China, in 2009 and the M.Sc. degree in Signal Processing from Nanyang Technological University, Singapore, in 2012, respectively. He worked as a Research Fellow with Future Mobility Research Lab, and Energy Research Institute @ NTU (ERI@N), Singapore. He is currently a Research Assistant Professor with the Department of Industrial Systems Engineering and Management, National University of Singapore, Singapore. His research interests focus on applying deep (reinforcement) learning to solve combinatorial optimization problems.
\end{IEEEbiography}

\begin{IEEEbiography}[{\includegraphics[width=1.0in, height=1.2in, clip,keepaspectratio]{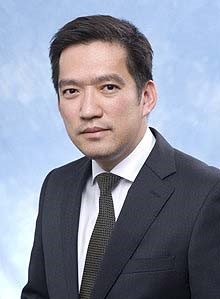}}]{Andrew Lim} received the Ph.D. degree in computer science in 1992 from the University of Minnesota, Minneapolis. He is currently a Professor with the department of Industrial Systems Engineering and Management, National University of Singapore (NUS). His works have been published in key journals such as Operations Research and Management Science, and disseminated via international conferences and professional seminars. Before Andrew was recruited by NUS under The National Research Foundation’s Returning Singaporean Scientists Scheme in 2016, he spent more than a decade in Hong Kong where he held professorships in The Hong Kong University of Science and Technology and City University of Hong Kong.
\end{IEEEbiography}

\begin{IEEEbiography}[{\includegraphics[width=1.0in, height=1.2in, clip,keepaspectratio]{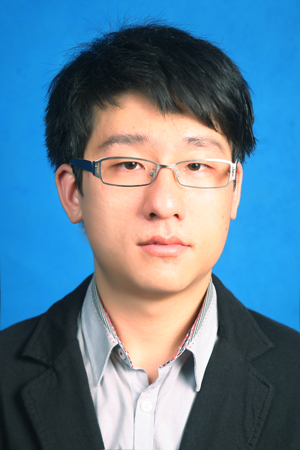}}]{Wen Song} {Wen Song} received the B.S. degree in automation and the M.S. degree in control science and engineering from Shandong University, China, in 2011 and 2014, respectively, and the Ph.D. degree in computer science from the Nanyang Technological University, Singapore, in 2018. He was a Research Fellow in the Singtel Cognitive and Artificial Intelligence Lab for Enterprises (SCALE@NTU). He is currently an Associate Professor with the Institute of Marine Science and Technology, Shandong University, China.

His current research interests include artificial intelligence, planning and scheduling, multi-agent systems, and operations research.
\end{IEEEbiography}

\begin{IEEEbiography}[{\includegraphics[width=1.0in, height=1.2in, clip,keepaspectratio]{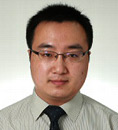}}]{Jie Zhang} received the Ph.D. degree from Cheriton School of Computer Science, University of Waterloo, Canada, in 2009. He is an Associate Professor with the School of Computer Science and Engineering, Nanyang Technological University, Singapore. During his Ph.D. study, he received the prestigious NSERC Alexander Graham Bell Canada Graduate Scholarship for top Ph.D. students across Canada. His research has been focused on the design of effective and robust intelligent software agents, through the modeling (trustworthiness, preferences) and simulation of different agents in a wide range of environments, using AI techniques (data mining, machine learning, and probabilistic reasoning) and multi-agent technologies.
\end{IEEEbiography}

\end{document}